\documentclass[conference]{IEEEtran}
\IEEEoverridecommandlockouts
\usepackage{cite}
\usepackage{amsmath,amssymb,amsfonts}
\usepackage{algorithmic}
\usepackage{graphicx}
\usepackage{textcomp}
\usepackage{xcolor}
\def\BibTeX{{\rm B\kern-.05em{\sc i\kern-.025em b}\kern-.08em
    T\kern-.1667em\lower.7ex\hbox{E}\kern-.125emX}}

\makeatletter
\def\ps@IEEEtitlepagestyle{%
  \def\@oddhead{}\def\@evenhead{}%
  \def\@oddfoot{%
    \hfil
    \begingroup
      \setlength{\fboxsep}{6pt}
      \setlength{\fboxrule}{0.4pt}
      \fbox{\parbox{0.98\textwidth}{\footnotesize
      This version of the manuscript is the original author-submitted version to \textit{IEEE International Symposium on Technology and Society (ISTAS 2025)}. Please cite the final version when available on IEEE Xplore.}}
    \endgroup
    \hfil
  }%
  \let\@evenfoot\@oddfoot
}
\makeatother

\begin{document}

\title{Human-AI Use Patterns for Decision-Making in Disaster Scenarios: A Systematic Review\\
\thanks{This work was supported by the AI Research Institutes Program funded by the National Science Foundation under the NSF AI Institute for Societal Decision Making (AI-SDM), Award No. 2229881}
}

\author{\IEEEauthorblockN{Emmanuel Adjei Domfeh}
\IEEEauthorblockA{\textit{Department of Computer Science and Engineering} \\
\textit{The Pennsylvania State University}\\
University Park, PA, USA \\
eza5399@psu.edu}
\and
\IEEEauthorblockN{Christopher L. Dancy}
\IEEEauthorblockA{\textit{Department of Industrial and Manufacturing Engineering} \\
\textit{The Pennsylvania State University}\\
University Park, PA, USA\\
cdancy@psu.edu}
\and
}

\maketitle
\thispagestyle{IEEEtitlepagestyle}

\begin{abstract}
In high-stakes disaster scenarios, timely and informed decision-making is critical—yet often challenged by uncertainty, dynamic environments, and limited resources. This paper presents a systematic review of Human-AI collaboration patterns that support decision-making across all disaster management phases. Drawing from 51 peer-reviewed studies, we identify four major categories: Human-AI Decision Support Systems, Task and Resource Coordination, Trust and Transparency, and Simulation and Training. Within these, we analyze sub-patterns such as cognitive-augmented intelligence, multi-agent coordination, explainable AI, and virtual training environments. Our review highlights how AI systems may enhance situational awareness, improves response efficiency, and support complex decision-making, while also surfacing critical limitations in scalability, interpretability, and system interoperability. We conclude by outlining key challenges and future research directions, emphasizing the need for adaptive, trustworthy, and context-aware Human-AI systems to improve disaster resilience and equitable recovery outcomes.
\end{abstract}

\begin{IEEEkeywords}
Human-AI Collaboration, Disaster Management, Decision Support Systems, Multi-Agent Coordination, Explainable Artificial Intelligence (XAI), Trust and Transparency, Simulation and Training
\end{IEEEkeywords}

\section{Introduction}
Decision-Making in high stake scenarios like disasters typically involves complex variables, limited resources, unpredictable conditions etc. In such cases, one of the most important objectives is to act within the shortest possible time. However, human responders are often overwhelmed by the urgency of tasks ranging from situational assessment and resource allocation, and these scenarios can be thought of as involving multi-criteria decision making which is of much interest to the general research community. The increased implementation and use of AI and related systems presents a dossier of opportunities that could be applied in solving cases of such scenarios.
Whether in processing vast amounts of data in real-time, predicting outcomes, and automating repetitive tasks, AI systems have promising potential for high stake scenarios. Some of the applications of AI systems to disaster management include easy navigation to safe points during disasters \cite{refPriyadarshi}, edge analytic decision-making systems for aerial vision processing \cite{refWagner}, IoT-enhanced AI for resource optimization \cite{refPraveen}, supervisory decision-making in emergency response \cite{refDilmaghani}, damage assessment via aerial imaging \cite{refMurphy}, and a host of others. However, integration of AI systems into disaster response and management scenarios requires thoughtful design that will ensure a considerate alignment to human needs, trust and adaptability. 
The concept of Human-AI use patterns provides a manner in which to better understand the most effective ways for humans to interact with AI systems given a context. During our review, we found several categories of patterns during disaster response that include collaborative learning, task coordination, resource management, and a host of others that are discussed in detail in this paper. By synthesizing insights from existing literature, this review aims to inform the design of scalable and equitable Human-AI systems tailored for disaster management.

\subsection{Human-AI Use Patterns }
Human-AI use patterns may be defined as a structured coordination of interactions and collaborative processes that occur between AI and humans. These patterns can be used as a conceptual tool to improve task performance, decision-making, and problem solving across many domains. They also can be used to explain how humans may share sub-tasks with and give information to AI systems in ways that allow mutual adaptation and achieving particular goals. Relatedly, Biloborodova \& Skarga-Bandurova\cite{refBiloborodova} see Human-AI collaboration, as inspired by the need to correct human fallibility and AI errors in decision-making. 

With the aim of proposing a compact generalization of existing practices and support the creation of new systems via expanding the possibilities of Human-AI interactions, Tsiakas and Murray-Rust \cite{refTsiakas} suggested a semi-formal design space for Human-AI interactions. They focused on interaction primitives that stresses a particular communication between humans and AI systems. They further showed how these primitives could be integrated into patterns providing an abstract specification for message exchanges between humans and AI models to facilitate purposeful interactions. Similar to their work, Gomez et al. \cite{refGomez} presented a taxonomy of interaction patterns in AI collaborative decision-making. They reported that the current outlook of human AI patterns is dominated by simplistic paradigms and as a result recommended the need for interactive functionalities to achieve effective human-AI collaboration, aiming for clear communication, trustworthiness, and enhanced decision-making processes. 

\subsection{Human-AI use patterns in disaster scenarios }

The application of Human-AI collaboration in disaster scenarios has seen a lot of development. Most of this research is aimed at improving the efficiency and effectiveness of emergency response efforts. The latter part of this paper delves more into the various classifications and patterns observed. In this part however, we cover briefly, human-AI use patterns in disaster scenarios.

A popular application area in Human-AI systems is AI-Based Drone Assistance in Human Rescue. Papyan et al.\cite{refPapyan} describes the use of AI-powered drones equipped with auditory detection capabilities that could identify human stress signals like screams in disaster scenarios. The advantage it offers is its ability to maneuver areas that may be inaccessible to human rescuers thereby accelerating search and rescue operations. 
 Papaioannou et al.\cite{refPapaioannou} focused on synergizing human responses and machine learning for disaster planning. They do so by proposing an attention-based cognitive architecture inspired by the Dual Process Theory. This framework integrates rapid, heuristic human-like responses with the optimized planning capabilities of AI, enabling autonomous agents to make complex decisions in dynamic disaster scenarios. 

In the area of large language models, Goecks and Waytowich\cite{refGoecks} introduced DisasterResponseGPT to swiftly generate actionable plans in the event of disaster response. They incorporated disaster response guidelines into the LLM and effectively generated multiple plans within a short period of time facilitating real-time decision-making and adaptability during emergencies. While these works demonstrate potential for reducing response times during disaster scenarios, they also raise important concerns about overreliance on AI technologies. The integration of AI with human expertise must be approached cautiously, as system failures, biases in training data, and lack of transparency can undermine trust and effectiveness in high-stakes situations. Although collaboration between human responders and AI systems shows promise, it remains essential to rigorously evaluate these systems in real-world conditions and ensure that human judgment remains central in complex emergency management.

\section{Methods}
To understand the current state of art of Human AI use patterns in disaster scenarios, we delved through three major journal databases with a curated search  to retrieve relevant papers. In this section of the paper, we discuss the underlying methodology for this literature review. By so doing we explore the search strings, the databases considered, the inclusion and exclusion criteria and finally a PRISMA diagram summary of the selected literature as shown in Figure 1. 

The search strings in this paper were motivated by the research theme as we intended to explore the current state of Human AI use patterns in disaster scenarios. We also looked at the critical application areas that could be interrogated to establish categories and classifications in this critical research area. The search strings were repeated in the following journal databases to retrieve relevant papers: ACM Digital Library, IEEE Xplore, and Google Scholar. The search strings used among others included: “Human-AI Collaborations”, “Human-AI use cases”, “Applications of Human AI use cases in Disaster Scenarios”, “Patterns in Human-AI use cases in Disaster”, Artificial intelligence in emergency response”, “AI for real-time disaster response” and other similar strings.
 
\subsection{Inclusion Criteria}
This study applied specific inclusion criteria to select literature focused on Human-AI collaboration in disaster recovery. Priority was given to peer-reviewed studies with empirical evidence, novel approaches, and relevance to decision-making, trust, and ethics in emergency response contexts. 

\begin{enumerate}
    \item \textbf{Focus on Disaster Scenarios:} Papers must explicitly address Human-AI collaboration in disaster recovery contexts, including emergency response, disaster planning, or post-disaster recovery.
    \item \textbf{Empirical Evidence:} Studies that demonstrate practical applications or empirical evaluations of Human-AI use patterns in disaster scenarios.
    \item \textbf{Novel Contributions:} Research presenting new Human-AI use patterns, frameworks, or methodologies tailored for disaster management. This sharpens our search for patterns that are relevant to the research study.
    \item \textbf{Relevance to Decision-Making: }Papers must explore Human-AI collaboration in decision-making processes during disaster scenarios. This is central to the underlying theme of this research.
    \item \textbf{Published in Peer-Reviewed Journals or Conferences:} Only studies from credible and peer-reviewed sources are included. This motivated the selection of papers from ACM Digital Library, IEEE Xplore, and Google Scholar.
    \item \textbf{Cross-Dimensional Analysis:} Papers that addressed trust, transparency, resource optimization, and ethical considerations in Human-AI collaboration were considered
\end{enumerate}

\subsection{Exclusion Criteria}
To maintain the relevance and quality of the review, certain studies were excluded based on the following criteria, which ensured that only rigorous and contextually appropriate research was considered. 
\begin{enumerate}
    \item \textbf{Lack of Disaster Context:} Papers not explicitly linked to disaster recovery, response, or planning were discarded.
    \item \textbf{Theoretical Discussions Only:} Studies that lack practical or empirical evidence for the proposed Human-AI use patterns. This is hinged on the authenticity and standardization of the research material considered.
    \item \textbf{Outdated or Redundant Research:} Studies published significantly earlier and superseded by recent advancements in the field. That is more than 10 years
    \item \textbf{Limited Scope:} Papers addressing only narrow aspects of Human-AI collaboration without broader implications for disaster scenarios.
    \item \textbf{Methodological Flaws:} Studies lacking rigor in methodology or having unclear research outcomes.
\end{enumerate}

\begin{figure}[htbp]
\centerline{\includegraphics[width=\linewidth]{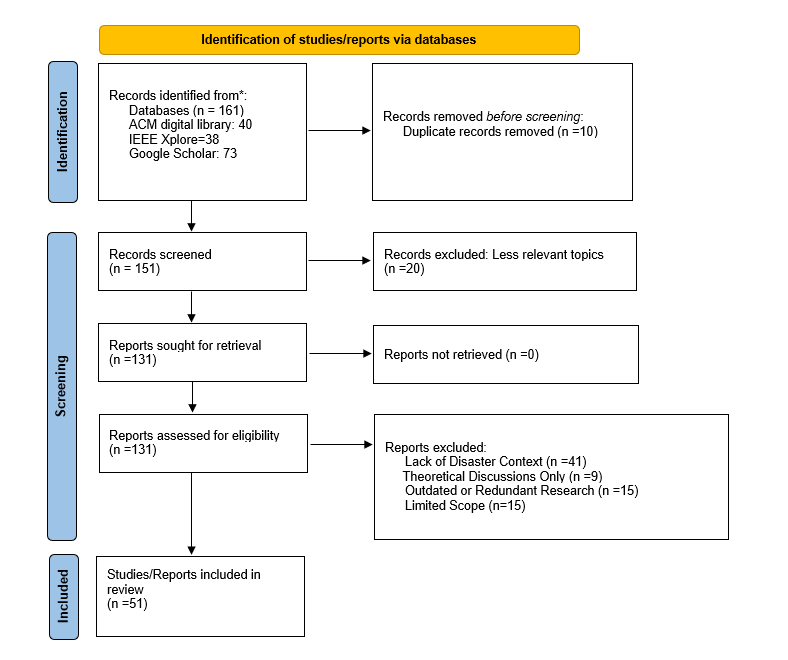}}
\caption{PRISMA flow diagram of the study selection process}
\label{fig:prisma}
\end{figure}

\section{Key Patterns and Findings}

In coherence with the research aim of identifying prominent human AI use patterns and the relevant techniques there in, this section highlights and discusses the patterns and key findings from the literature reviews. 
Figure 2 shows the categorizations of the key findings seen in our review and the subsequent subsections explains them further. It is worth-noting that some papers fell within multiple categories because of their reach. 

\begin{figure}[htbp]
\centerline{\includegraphics[width=\linewidth]{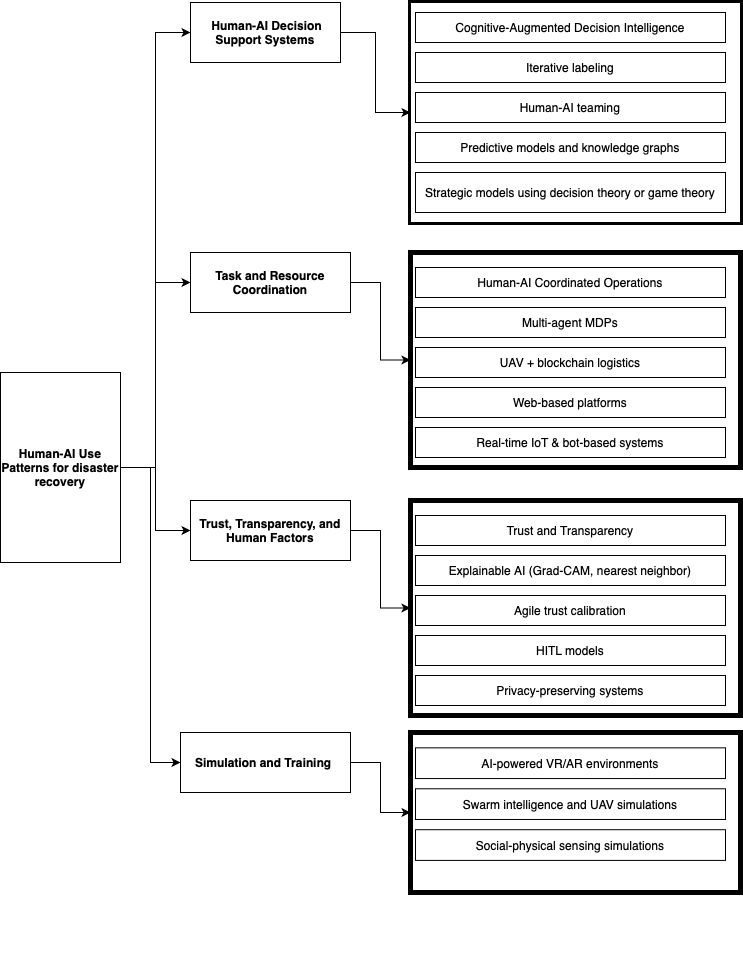}}
\caption{Summary of categorized patterns in Human AI use patterns}
\label{fig:categories}
\end{figure}

\subsection{Human-AI Decision Support Systems}

Disaster scenarios are characterized by uncertainty, time-critical decisions, and information overload. Human-AI Decision Support Systems enable real-time, informed decision-making by integrating algorithmic intelligence with human cognition. These systems bridge data and decisions by leveraging a range of strategies as outlined below:

\subsubsection{Cognitive-Augmented Decision Intelligence}

The AIDR platform exemplifies cognitive augmentation by integrating crowd-sourced human labeling and machine learning to process crisis data from social media platforms such as Twitter \cite{refImran}. The system improves classification of disaster-relevant content, demonstrating how AI can augment human sense-making during chaotic situations. A similar approach by Fan et al.\cite{refFan} proposed a “Disaster City Digital Twin” that combines machine intelligence with human expertise for continuous monitoring and situational modeling, representing a scalable cognitive augmentation strategy for urban crisis planning. 

Beyond textual and structured data, cognitive augmentation has also been explored through multimodal fusion techniques. Pouyanfar et al.\cite{refPouyanfar} developed a deep learning framework that processes disaster-related video and audio streams using separate neural networks and fuses them via multiple correspondence analysis (MCA) to improve crisis insight and classification accuracy. Despite limited direct human-in-the-loop interaction, the system still depends on human-supervised data curation and evaluation, reinforcing the importance of human cognition in guiding multimodal AI for decision support. Their approach achieved up to 73\% accuracy on YouTube disaster datasets, demonstrating the potential of multimodal models to enhance situational awareness in complex disaster environments.

Mitigating cognitive biases, particularly anchoring, has been identified as critical in human--AI decision-making. Rastogi et al.\cite{rastogi2022deciding} developed a biased Bayesian framework to formally model cognitive biases and conducted two user experiments to test mitigation strategies. Their results showed that allocating more time enabled participants to adjust away from incorrect AI predictions, with disagreement rates rising from 48\% at 10 seconds to 67\% at 25 seconds. Building on this, they proposed a confidence-based time allocation policy, giving more time when AI confidence is low and less when confidence is high, which, when combined with explanations, further reduced anchoring bias and improved human--AI collaborative accuracy.

 Abid et al.\cite{refAbid} illustrated how AI may improve disaster management using technologies such as machine learning, deep learning, and convolutional neural networks to predict hazards like earthquakes, landslides, and floods. Tools such as MOBILISE integrate GIS with AI to offer real-time situational awareness and risk visualizations that support effective decision-making and resource allocation. LSTM models can be used process diverse data—from satellite imagery to social media text—for tasks ranging from flood zoning to real-time damage assessment. These findings indicate that AI may contribute to scalable, data-informed approaches to disaster response and recovery, provided its implementation is context-aware and carefully evaluated.

Another relevant contribution in this space is the conceptual framework proposed by Molina et al. \cite{refMolina}, which applies the Extended Mind Theory to disaster decision-making contexts. The framework combines AI’s capacity for data processing, pattern recognition, and predictive analysis with expert human judgment to enhance organizational decision-making, especially in dynamic and unstructured environments. It emphasizes balancing the data-processing power of AI systems with human creativity to enhance decision accuracy and speed. Although primarily conceptual and lacking extensive empirical validation, the model raises important considerations around aligning AI outputs with human cognitive models, ensuring transparency, and mitigating algorithmic bias. The work highlights the critical need for rigorous evaluation in real-world settings while advancing the conversation around cognitive augmentation in complex decision environments.

\subsubsection{Iterative Labeling}

AIDR also leverages active learning in which human annotators label a small subset of tweets, which are used iteratively by the model to enhance accuracy. This human-in-the-loop approach enables dynamic adaptation to evolving disaster contexts. The system achieved 80 percent AUC in the Pakistan Earthquake response, highlighting its practical value \cite{refImran}. However, reliance on continuous crowdsourcing introduces scalability issues and annotation fatigue. The broader framework of “AI for Good” \cite{refKshirsagar} further reinforces iterative labeling by engaging NGO experts to iteratively refine models in data-scarce environments, using metadata-driven augmentation techniques.

\subsubsection{Human-AI Teaming}

DisasterResponseGPT illustrates effective teaming by embedding domain-specific response protocols into a large language model, allowing it to generate actionable disaster response plans within seconds \cite{refGoecks}. The system outputs structured recommendations for human validation, enabling collaborative decision-making. A similar teaming effort was discussed in  Molina et al.\cite{refMolina}, who emphasize conceptual frameworks for effective collaboration and aligned decision logic in real-world deployments.

\subsubsection{Predictive Models and Knowledge Graphs}

Some systems utilize structured data representations such as knowledge graphs and semantic modeling for forecasting and prioritization. For instance, Yang et al. \cite{refYang} proposed a knowledge graph-based system for emergency coal mine management, improving the semantic accuracy of situation updates and enabling pattern recognition. Additionally, Comes \cite{refComes} outlines conceptual best practices in predictive modeling for crisis decision systems.

Reddy et al.\cite{refReddy} proposed RescueMe, an AI-powered decision support framework integrating real-time data analysis, predictive modeling, and geolocation to enhance disaster response. Leveraging NLP, machine learning, and satellite/social media data, the system predicts disasters, maps evacuation routes, and supports post-disaster recovery through dynamic data visualization. Its chatbot capabilities also facilitate real-time communication with affected individuals, improving coordination among emergency responders and stakeholders.

Linardos et al.\ \cite{refLinardos} reviewed ML/DL methods in disaster management across all phases—from early warnings to post-disaster recovery—using models like CNNs and LSTMs on satellite and social media data. Their results showed high accuracy (e.g., 99\% for wildfire detection), with DL models excelling in unstructured data processing and ML in mitigation. They advocate for integrating explainable AI with big and crowdsourced data while addressing challenges like data quality and long-term recovery modeling.

Tao et al.\ \cite{refTao} emphasized the role of predictive analytics and data visualization in pandemic decision-making. Their survey showed that data-driven methods improved preparedness and response accuracy, though real-time data access and infrastructure posed practical constraints.

Cao \cite{refCao} introduced the Smart Disaster Resilience (SDR) framework, expanding the traditional MPRR model with AI-driven capabilities such as early warning, forecasting, and multi-hazard management. By integrating diverse data from environmental, social, and cyber systems, the framework enables adaptive, real-time decision-making. The study emphasizes the need for ethical, explainable, and user-centric AI to ensure stakeholder trust and broad adoption in intelligent disaster resilience strategies.

\subsubsection{Strategic Models Using Decision or Game Theory}

In complex, multi-stakeholder environments, strategic AI models using decision-theoretic or game-theoretic reasoning optimize path planning and resource coordination. Papaioannou et al. \cite{refPapaioannou} presented a cognitive architecture that combines heuristic human-like decision processes with attention-based planning mechanisms. These models emulate both autonomous and assisted strategies during rapidly evolving disaster scenarios.

Sun et al.\cite{refSun} examined supervised, unsupervised, and reinforcement learning models supporting all phases of disaster management. Their study showed that such AI systems can enhance resource allocation and recovery planning, reducing human error in high-stakes decisions. However, they also noted computational limitations that may hinder broader adoption.

Human-AI Decision Support Systems can contribute significantly to reducing cognitive load, accelerating time-to-decision, and enhancing the precision of situational assessments in disaster scenarios. By combining data-driven foresight with human contextual awareness, these systems enable adaptive and scalable interventions. However, their efficacy is often constrained by limited labeled datasets, heavy dependence on expert input, and generalizability across disaster types. Challenges also persist around model explainability, which is essential for trust-building and real-world adoption.

\subsection{Task and Resource Coordination}

In disaster recovery, timely allocation of resources, personnel, and technology is paramount. Task and Resource Coordination addresses this by integrating human judgment with AI-driven automation to dynamically assign tasks, prioritize needs, and optimize logistics under uncertainty.

\subsubsection{Human-AI Coordinated Operations}
Many systems embrace decentralized coordination, combining human situational awareness with automated AI agents. For example, Papaioannou et al. \cite{refPapaioannou} proposed an attention-based hybrid planning model that merges heuristic decision-making with AI scheduling to enhance coordination in volatile environments. These models simulate both cognitive and operational decisions, offering resilience under uncertainty. Al-hussaini\cite{refAl-hussaini} proposed an Intelligent Decision Support System (IDSS) tailored for multi-robot coordination in disaster scenarios, designed to function effectively even with delayed or incomplete information. By combining heuristic task selection with probabilistic reasoning, the system enhances the decision-making speed and reliability of human supervisors. Its task reassignment capabilities further support operational continuity by mitigating risks and improving coordination in dynamic, communication-constrained environments.

Dilmaghani and Rao\cite{refDilmaghani} explored supervisory decision-making frameworks to address communication and workflow bottlenecks in emergency response systems. They employed discrete event system modeling to simulate and predict operational disruptions, allowing for timely supervisory interventions that maintain system flow and prevent deadlocks. This high-level coordination enhances the efficiency of Human-AI collaborations during time-critical disaster operations.

Zhang\cite{refZhang} examined how artificial intelligence technologies can enhance emergency logistics by enabling demand forecasting, accurate supply--demand matching, rapid circulation of emergency materials, and real-time supervision of logistics flows. The study emphasized that AI can address problems such as shortages of critical supplies, mismatches between donated goods and actual needs, and transportation delays caused by restrictions during public health emergencies. Rather than presenting quantitative performance results, the paper contributes a conceptual model that integrates AI into forecasting, warehousing, unmanned distribution, and inventory management to improve the responsiveness and efficiency of emergency logistics systems. Similarly, Afkhamiaghda and Elwakil\cite{refAfkhamiaghda} proposed a KNN-based classification model to optimize temporary housing allocation based on socio-demographic and environmental variables. This model achieved 85\% classification accuracy, supporting efficient, data-informed decisions that matched population needs with available shelter resources.

\subsubsection{Multi-agent MDPs}
Stochastic planning under uncertainty is often managed through multi-agent Markov Decision Processes (MDPs). Ramchurn et al.\cite{refRamchurn} modeled human-agent collaboration using such frameworks to simulate collaborative decision-making for disaster relief, such as resource deployment in dynamically changing conditions. With the use of their AtomicOrchid simulation, they affirm the value of consolidating human decision-making with propositions by AI agents. This enables commanders to adopt supervisory roles and flexibly respond to dynamic changes during disaster.  Similarly, Abeywickrama et al. \cite{refAbeywickrama} used formal model-checking approaches to verify responsible behavior in multi-agent coordination settings. Their model proposition is suited for multi-agent scenarios while ensuring safety, controllability, and ethical behavior in human-AI interactions. This contribution also makes it part of the Trust and Transparency category of the patterns identified.

\subsubsection{UAV + Blockchain Logistics}
AI-powered UAVs equipped with blockchain for secure and traceable supply chain operationsalso appear in the literature. Papyan et al.\cite{refPapyan} demonstrated how AI-based drone systems could detect human stress cues in inaccessible zones, while \cite{refWang} introduced the RescueChain framework combining blockchain, vehicular fog computing, and deep reinforcement learning to improve offloading and data transmission efficiency during emergencies.
Vedanth et al. \cite{refVedanth} introduced the use of YOLOv8-integrated AI drones to autonomously detect disaster victims and perform supply distribution. They present an improved optimal configuration that improved detection accuracy significantly compared to default parameters.
Similarly, Allen and Mazumder \cite{refAllen} proposed the AASAPS-HADR system, which integrates UAVs and AI for aerial surveys to assess damage, locate survivors, and coordinate emergency routing. Their experiments provided direct, empirical proof that AI-based perception (pose recognition + frequency analysis) can be executed onboard small UAVs to support disaster triage 

\subsubsection{Web-based Platforms}
Cloud-enabled dashboards and web-based coordination tools allow multiple stakeholders to manage task assignment, logistics, and volunteer coordination in real time. Hanji et al.\cite{refHanji} presented a disaster management web application that harmonizes data streams and resource allocations in crisis zones through a shared digital interface, even in areas with unstable connectivity \cite{refAllen}.

\subsubsection{Real-time IoT and Bot-based Systems}
IoT sensors and bot-based systems support real-time decision support through environmental monitoring and automation. \cite{refGladence} proposed a swarm intelligence model using bot agents to coordinate response tasks and manage decentralized communication in disaster scenarios. 

Kanth et al\cite{refkanth} introduced Disastro, a Twitter-based disaster response system that classifies real-time social media posts into actionable categories such as "rescue" and "donation". This automated classification enables emergency responders to prioritize interventions amidst high-velocity data streams, particularly during dynamic environmental conditions. Similarly, Mondal et al.\cite{refMondal} proposed the use of information bots within mobile opportunistic networks to maintain data accuracy and communication continuity when traditional infrastructure fails. Their system achieved 85\% classification accuracy and significantly improved task coordination by validating and disseminating critical disaster information across affected regions.

Qadir et al.\ \cite{refQadir} present a comprehensive view of crisis analytics by applying big data techniques to disaster phases—preparedness, response, and recovery. By leveraging mobile data, social media, crowdsourced platforms, and IoT sensors, they demonstrate how AI-driven systems enhance population tracking, resource allocation, and decision-making in real time. Their work highlights practical applications in various global crises, while addressing concerns of privacy, digital inclusion, and bias mitigation through human-AI hybrid models.

\subsubsection{Simulation Models for Coordination}
Digital twins, such as the “Disaster City Digital Twin” \cite{refFan}, demonstrate how simulating interdependencies across resources, agents, and geography can improve resource allocation decisions. These simulations enable proactive task prioritization and distributed planning in advance of real-world deployment.

\textbf{Contributions and Challenges.}  
Task and Resource Coordination models improve operational agility, enhance scalability across geographies, and reduce response latency. However, challenges persist in integrating heterogeneous systems, ensuring low-latency communication, and calibrating trust among multi-agent systems. Moreover, ensuring human oversight in AI-automated planning loops and modeling agent interdependencies remain key research frontiers in building adaptive and resilient coordination platforms.

\subsection{Trust, Transparency, and Human Factors}

In disaster scenarios where decisions impact lives, fostering trust and interpretability in AI systems is critical. This category includes Human-in-the-Loop (HITL) models, explainable AI techniques, and frameworks that promote ethical alignment and user confidence in automated systems.

\subsubsection{Trust and Transparency}
Transparent interaction design is key to human-AI trust calibration. Biloborodova and Skarga-Bandurova\cite{refBiloborodova} emphasize that reliable disaster decision-making is contingent on confidence in system outputs. They propose trust models that incorporate feedback loops between human operators and AI agents to maintain system accountability and decision traceability. In critical settings like healthcare, similar models have shown how perceived reliability can influence adoption decisions \cite{refHemmer}.

Hemmer et al.\cite{refSrivastava} explored strategies to enhance shared mental models and situational awareness in human-AI teams. By improving transparency in the AI decision-making process, their study demonstrated that human operators could better understand system behavior, which reduced over-reliance and led to more accurate and contextually grounded decisions. This reinforces the importance of cognitive alignment and clarity in interface design when developing trustworthy AI tools for high-stakes environments.
\subsubsection{Explainable AI (Grad-CAM, nearest neighbor)}
Explainable AI (XAI) enhances interpretability of predictions through visual or symbolic explanations. Humer et al.\cite{refHumer} explored how techniques such as Grad-CAM and nearest neighbor exemplars affect human decisions in safety-critical tasks. Their findings show that while explanations can build confidence, they may also mislead users if not aligned with their mental models.

\subsubsection{Agile Trust Calibration}
Okamura and Yamada\cite{refOkamura} introduced adaptive trust calibration mechanisms, where trust levels are dynamically adjusted based on observed AI performance and task difficulty. These models prevent over-reliance or under-utilization by modulating user engagement with system outputs, thus improving human-AI teaming in evolving disaster situations.

Van Den Bosch and Bronkhorst \cite{refBosch} emphasize the evolution of AI from basic assistants to adaptive teammates in high-stakes environments like military operations. Their study highlights the need for trust-aware and proactive AI systems that can enhance situational awareness, mitigate biases, and adapt to dynamic team contexts for more effective human-AI cooperation.

Hemmer et al. \cite{refHemmer} explored adaptive AI systems designed to enhance disaster planning by promoting mutual learning, user adaptability, and decision transparency. They identified six factors for adoption success—including complementarity, time efficiency, and explainability—arguing that well-integrated AI can improve trust, accuracy, and resilience during critical decision-making under stress.

\subsubsection{Human-in-the-Loop (HITL) Models}
Human-in-the-Loop (HITL) paradigms ensure that human expertise remains central in AI-supported decisions. Kshirsagar et al.\cite{refKshirsagar} argue for integrating domain experts into the feedback cycle to contextualize AI predictions and uphold ethical standards during rapid disaster responses. Rashid et al.\cite{refCaldwell} also emphasize HITL integration within agile research frameworks for trust-sensitive design.
In high-stake disaster scenarios, ensuring responsible AI behavior is equally vital. Abeywickrama et al.\cite{refAbeywickrama} advanced this notion by introducing a model-checking framework tailored for human-agent collectives. Their work utilizes the Model Checker for Multi-Agent Systems (MCMAS) to verify essential properties such as safety, ethical soundness, and controllability in UAV-based disaster response. This verification mechanism strengthens accountability and ensures that AI decisions remain transparent, ethical, and aligned with human oversight mandates.

\subsubsection{Privacy-Preserving Systems}
Privacy remains a critical concern in AI-powered disaster management, especially with sensitive geolocation, identity, or health data. Rashid et al.\cite{refRashid} and Gentile et al.\cite{refGentile} propose privacy-aware architectures that balance data utility with legal and ethical compliance. These include federated sensing models and voice-based interaction platforms that maintain confidentiality while aiding disaster recovery.

\textbf{Contributions and Challenges.}  
This category advances disaster AI by embedding ethical safeguards and making black-box models more understandable to end-users. It contributes frameworks for building trust, maintaining accountability, and preserving privacy in crisis contexts. However, challenges persist in balancing model transparency with performance, calibrating trust without inducing over-reliance, and ensuring privacy under time-critical conditions. Future work should prioritize real-world deployments and longitudinal studies to assess sustained trustworthiness and usability across diverse user groups.

\subsection{Simulation and Training}

Simulation and training frameworks serve as critical environments for testing, refining, and preparing both AI systems and human responders for real-world disaster recovery. By replicating complex disaster dynamics, these systems allow iterative learning, stress-testing of algorithms, and enhanced human-AI coordination in controlled yet realistic settings.

\subsubsection{AI-powered VR/AR Environments}
Immersive Virtual Reality (VR) and Augmented Reality (AR) platforms enable responders to engage with AI agents in disaster simulations prior to real-world deployment. Hemmer et al.\cite{refHemmer} argue that AI-augmented VR systems enhance mental model alignment and preparedness, allowing users to better anticipate AI behavior and optimize response planning.
Al-Rajab et al. \cite{refMurad} propose an AI-integrated platform aimed at real-time disaster management and volunteer training. By integrating historical data with machine learning algorithms, the system generated dynamic disaster response plans while leveraging a centralized database for resource optimization. VR volunteer training platform 85\% accuracy in generating response plans,  90\% effectiveness in volunteer matching 

Another simulation approach was introduced by Huang et al.\cite{refHuang}, who developed a UAV-based decision-making and command platform for quasi-real-time 3D mapping and situational awareness. Their system simulated disaster scenarios by integrating fog, cloud, and edge computing to visualize crisis zones and coordinate rescue teams. The platform included spatial analysis, route planning, and multi-source data fusion, forming a digital command structure that adapted to dynamic conditions. Results showed a 25\% reduction in decision-making delays and a 40\% increase in search-and-rescue efficiency, underscoring its real-world potential for disaster response.

Wagner and Roopaei\cite{refWagner} proposed a virtual simulation framework that leverages simulated environments to generate synthetic disaster scenarios for developing and testing decision-making systems. Their framework produces annotated aerial datasets that support training lightweight neural networks such as YOLOv3-Tiny for object detection and classification. These models are embedded within observational agents, such as UAVs, enabling them to autonomously detect and localize objects and provide situational updates in real time. The study emphasizes the feasibility of performing these tasks at the edge using resource-constrained devices and highlights the role of the virtual framework as a safe and cost-effective platform for preparing AI systems before real-world deployment.

Tsai et al.\cite{refTsai} focused on simulation and training through participatory design with tribal governments to co-develop AI-driven disaster response tools, including chatbots and drones. Their chatbot enables damage reporting via text, voice, and images, improving documentation for aid applications, while AI-integrated drones support wildfire control and water quality monitoring. The study highlights culturally sensitive, scalable disaster systems that enhance communication, preparedness, and resilience in tribal communities.

Wilchek\cite{refWilchek} introduced Ajna, a collaborative AR system that supports emergency sensemaking by combining AI-driven object detection with shared spatial awareness across multi-level environments. Components like SpatialSense and PercepShare enable teams to locate people and objects through walls, reducing search-and-rescue task time by 15\%. While Ajna boosts collaboration and situational efficiency, challenges like IMU drift and outdoor limitations remain.

\subsubsection{Swarm Intelligence and UAV Simulations}
Swarm intelligence principles underpin the simulation of distributed unmanned aerial vehicles (UAVs) for coordinated disaster relief operations. Gladence et al.\cite{refGladence} presented a UAV framework leveraging swarm heuristics for adaptive rescue in inaccessible zones, offering valuable training grounds for both autonomous agents and human operators.

\subsubsection{Social-Physical Sensing Simulations}
Hybrid simulation models incorporating social and physical data are emerging as testbeds for complex decision-making. Rashid et al.\cite{refRashid} introduced the concept of social-physical sensing systems that simulate collective intelligence from human-machine interactions—providing insight into communication dynamics, mobility patterns, and infrastructure status in disaster zones.

Fan et al.\cite{refFan} proposed the “Disaster City Digital Twin,” a visionary simulation framework that integrates artificial intelligence (AI) and information and communication technologies (ICT) to replicate and analyze disaster dynamics for improved planning and response. The framework is built on four components: multi-data sensing, data integration and analytics, multi-actor game-theoretic decision making, and dynamic network analysis. It enables visual monitoring, predictive modeling, and coordination among diverse stakeholders, while also supporting “what-if” scenario testing through serious gaming environments. 

\textbf{Contributions and Challenges.} 
Simulation platforms contribute to safer disaster preparation by enabling iterative feedback between AI and human agents in lifelike environments. They also offer the potential to generalize across different scenarios and train responders in decision-making under stress. However, challenges include ensuring realism in agent behavior, modeling unpredictable human responses, and achieving computational scalability across diverse environments. Continuous updates from real-world disaster data will be essential to improve fidelity and relevance of these platforms.

\section{Summary of Key Findings}

This systematic review identified and analyzed 51 peer-reviewed studies, categorizing Human-AI use patterns across four main themes: \textbf{Decision Support Systems}, \textbf{Task and Resource Coordination}, \textbf{Trust, Transparency, and Human Factors}, and \textbf{Simulation and Training}. These categories emerged from a rigorous synthesis of empirical evidence, conceptual frameworks, and applied systems within disaster response contexts.

\textbf{Human-AI Decision Support Systems (DSS)} highlight the convergence of AI-driven prediction, multimodal fusion, and cognitive augmentation with human expertise. Systems like AIDR and DisasterResponseGPT demonstrate iterative labeling and real-time crisis planning, while decision-theoretic models and knowledge graph-based forecasting tools underscore AI’s ability to enhance situational awareness, reduce cognitive load, and accelerate decision-making under pressure. However, challenges such as annotation scalability, transparency, and alignment with human cognition persist.

\textbf{Task and Resource Coordination} emphasizes the optimization of logistics and personnel through AI-integrated planning, UAVs, and blockchain-powered systems. Studies explored decentralized coordination using Multi-Agent MDPs, decision support for emergency logistics, and real-time classification via IoT and bot-based platforms. These systems show promise in improving operational agility, but scalability, system interoperability, and human oversight in automated loops remain unresolved.

\textbf{Trust, Transparency, and Human Factors} focus on building user confidence through explainable AI (XAI), adaptive trust calibration, HITL frameworks, and privacy-preserving models. Key contributions include visual explanation techniques (e.g., Grad-CAM), model-checking for ethical behavior in UAVs, and participatory AI design for transparent decision-making. While fostering trust, work in this domain grapples with balancing explainability and performance, mitigating over-reliance, and maintaining privacy under time-critical conditions.

\textbf{Simulation and Training} frameworks serve as lifelike testbeds for disaster readiness. VR/AR systems, swarm simulations, and digital twins like the "Disaster City Digital Twin" offer realistic environments for refining AI behavior and training human responders. These platforms support collaborative response planning, real-time data fusion, and scenario testing. However, limitations exist in ensuring model fidelity, simulating human behavior under stress, and scaling across diverse environments.

Collectively, these findings reveal that while AI can enhance speed, precision, and scalability in disaster recovery, responsible integration hinges on transparency, human-centered design, and cross-disciplinary collaboration.

\section{Future Directions and Challenges}

Despite notable advancements, several open challenges persist in operationalizing Human-AI collaboration for disaster management. First, \textbf{scalability and generalizability} remain limited—many systems are tested in narrow contexts with specific disaster types, making cross-domain deployment challenging. Future systems should adopt modular, transferable designs capable of adapting to dynamic disaster environments.

Second, \textbf{data quality and availability} continue to hinder model robustness. Real-time and multimodal data streams are often noisy, incomplete, or privacy-sensitive. Future research must prioritize robust preprocessing techniques, privacy-preserving data integration, and synthetic data generation for training and validation.

Third, \textbf{trust calibration and interpretability} are still emerging. While explainable AI (XAI) methods exist, their real-world efficacy under cognitive load and stress is not fully understood. Future work should evaluate XAI techniques in live deployments and adapt explanations based on user expertise and situational urgency.

Additionally, \textbf{human-in-the-loop system design} must evolve from static oversight models to adaptive teaming frameworks that dynamically assign control, support mutual learning, and minimize reliance without losing accountability. Integrating ethical reasoning, value alignment, and responsible AI practices is essential for sensitive, high-stakes scenarios.

Another future direction involves linking the value-sensitive framework in this paper with emerging human-AI interaction paradigms, such as those proposed by van Berkel et al. \cite{refVanBerkel}—intermittent, continuous, and proactive. These paradigms expand how AI systems engage users beyond traditional, turn-based interactions. In disaster settings, where decision-making is rapid and ongoing, continuous and proactive systems could better support human responders. Integrating the proposed use patterns into these modes offers a broader design space but also raises challenges around timing, user attention, and intervention relevance.

Finally, \textbf{interdisciplinary collaboration} must be strengthened. The integration of AI researchers, emergency managers, policy makers, and affected communities is critical for co-designing inclusive, context-aware, and sustainable disaster recovery systems. Future work should focus on longitudinal assessments, real-world deployments, and adaptive frameworks to ensure robustness across heterogeneous disaster scenarios.

\section{Conclusion}

This systematic review synthesized current literature on Human-AI use patterns in disaster scenarios, categorizing them into Decision Support Systems, Task and Resource Coordination, Trust and Transparency, and Simulation and Training. Across these categories, AI was shown to augment decision-making, enhance coordination, and facilitate realistic training environments. However, the deployment of these systems faces limitations in scalability, transparency, and empirical validation.

By aligning AI capabilities with human values, contextual expertise, and adaptive feedback, Human-AI systems can serve as robust allies in disaster management. The findings offer a roadmap for designing collaborative systems that not only automate but also empower, fostering inclusive and equitable response efforts. As disasters become more complex and frequent, the future of effective response lies in transparent, responsive, and human-centered AI systems.

\bibliographystyle{IEEEtran}
\bibliography{disaster_ai_references}

\end{document}